\documentclass[11pt,a4paper]{article}
\usepackage[hyperref]{acl2019}
\usepackage{times}
\usepackage{latexsym}

\usepackage{booktabs} 
\usepackage{graphicx}
\usepackage[utf8x]{inputenc}
\usepackage{amsmath}
\usepackage{multirow}
\usepackage{appendix}
\usepackage{url}
\usepackage{tikz}
\usepackage{makecell}
\usepackage{balance} 
\usepackage{flushend} 

\newcommand{\lden}{[\![}
\newcommand{\rden}{]\!]}
\newcommand{\denotes}[1]{\lden #1\rden}

\newcommand{\atc}[0]{denotative}
\newcommand{\cta}[0]{connotative}

\newcommand{\wac}[0]{\textsc{wac}}
\newcommand{\logreg}[0]{\textsc{lr}}
\newcommand{\mlp}[0]{\textsc{mlp}}
\newcommand{\tree}[0]{\textsc{dt}}

\usepackage{url}

\aclfinalcopy 

\title{Composing and Embedding the\\ Words-as-Classifiers Model of Grounded Semantics}

\author{Daniele Moro \\
  Boise State University \\
  1910 University Dr. \\
  Boise, ID 83725 \\
  \texttt{danielemoro@}\\ \texttt{u.boisestate.edu} \\\And
  Stacy Black \\
  Boise State University \\
  1910 University Dr. \\
  Boise, ID 83725 \\
  \texttt{stacyblack@}\\ \texttt{u.boisestate.edu} \\\And  
  Casey Kennington \\
  Boise State University \\
  1910 University Dr. \\
  Boise, ID 83725 \\
  \texttt{caseykennington@}\\ \texttt{boisestate.edu} \\}

\date{}

\begin{document}
\maketitle
\begin{abstract}
The words-as-classifiers model of grounded lexical semantics learns a semantic fitness score between physical entities and the words that are used to denote those entities. In this paper, we explore how such a model can incrementally perform composition and how the model can be unified with a distributional representation. For the latter, we leverage the classifier coefficients as an embedding. For composition, we leverage the underlying mechanics of three different classifier types (i.e., logistic regression, decision trees, and multi-layer perceptrons) to arrive at a several systematic approaches to composition unique to each classifier including both denotational and connotational methods of composition. We compare these approaches to each other and to prior work in a visual reference resolution task using the refCOCO dataset. Our results demonstrate the need to expand upon existing composition strategies and bring together grounded and distributional representations.
\end{abstract}

\section{Introduction}

The words-as-classifiers (\wac) model of lexical semantics has shown promise as a way to acquire grounded word meanings with limited training data in interactive settings. Introduced in \newcite{Kennington2015a} for grounding words to visual aspects of objects, \newcite{Schlangenetal2016} showed that the model could be generalized to work with any object representation (e.g., a layer in a convolutional neural network) with ``real'' objects depicted in photographs. The \wac\ model builds on prior work \cite{larsson2015formal} treating formal predicates as classifiers to effectively learn and determine class membership of entities (e.g., an object $x$ denoted as ``red'' in an utterance belongs to the predicate $red(x)$ class). The \wac\ model has been used to ground into modalities beyond just vision, including simulated robotic hand muscle activations \cite{Moro2018}, and  \wac\ has been used for language understanding in a fluid human-robot interaction ask \cite{hough-schlangen:2016:SIGDIAL} . Beyond comprehension tasks, the \wac\ model has also been used for referring expression generation \cite{zarriess2017child}. The \wac\ classifiers can use any set of features from any modality and the model is \emph{interpretable} because each word has its own classifier. Moreover, the \wac\ model allows for incremental, word-by-word composition which has implications for interactive dialogue: human users of incremental spoken dialogue systems perceive them as being more natural than non-incremental systems \citep{Aist2006,Skantze2009,Asri2014}.

Though the \wac\ model has pleasing theoretical properties and practical implications for interactive dialogue and robotic tasks, \newcite{Boyer2012} and \newcite{Emerson2017} point out that \wac\ treats all words independently, thereby ignoring distributional relations and meaning representations, and \emph{composition} using the \wac\ model has generally resulted in averaging over applications of \wac\ to objects--a purely intersectional approach to semantic composition, which has been shown to fail in many cases \cite{Kamp1975}. As is well known, linguistic structures are compositional---simple elements can be functionally combined into more complex elements \cite{Frege1892}---and composition, which is an important aspect of grounding into visual or other modalities, is itself a process that is a function of how a particular model of lexical semantics is represented.\footnote{To illustrate, following \newcite{baroni-zamparelli:2010:EMNLP}, the \emph{formal semantics} approach to lexical semantics, as derived from \newcite{montague1970universal,Montague1973a}, treats lexical meaning as predicates which functionally determine class membership. In contrast, \emph{distributional} representations (e.g., vector space models \cite{Turney2010} including embeddings) differ from formal semantics in that word meanings are represented not as predicates, but as high-dimensional vectors and as a result approaches to composition revolve around tensor operations from summation and multiplication \cite{Mitchell2010} to treating adjectives as matrices and nouns as vectors \cite{baroni-zamparelli:2010:EMNLP}.}

\begin{figure}[t]
  \centering
      \includegraphics[width=0.45\textwidth]{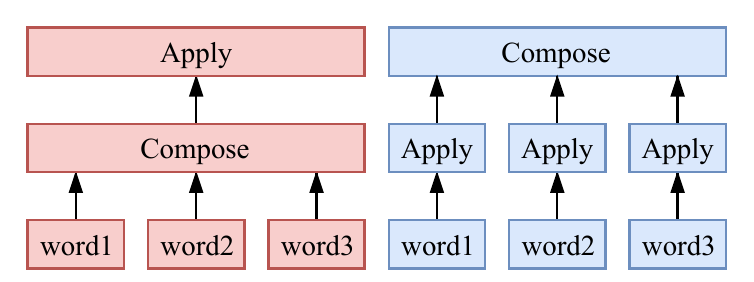}	
      \caption{Left: connotational \emph{compose-then-apply}; Right: denotational \emph{apply-then-compose}. \label{fig:composition-types}}
      \vspace{-5mm}
\end{figure}

The goal of this paper is (1) to explore and understand possible approaches to composition for \wac, and (2) by determining the best \wac\ model, we extend \wac\ by using classifier coefficients as embedding vectors in a semantic similarity task. At the outset, following \newcite{Schlangenetal2016}, we make an important distinction by identifying two composition types: in any task, composition can be handled at the level of \emph{connotation} where words are composed \emph{then} applied in the task (e.g., tensor operations of word embeddings are generally composed by summation before they are used in a task, such as sentiment classification), and at the level of \emph{denotation} where words are applied then composed in a task (e.g., \wac\ classifiers are applied to object representations, then the results of those applications are composed, such as reference resolution to visual objects). These differences in composition process are depicted in Figure~\ref{fig:composition-types} which distinguishes between application to a task and composition.

More specifically, we alter the \wac\ model to leverage several classifier types, namely logistic regression, multi-layer perceptrons, and decision trees which allow us to use both \cta\ and \atc\ composition strategies and benefit from the different classifiers, for example by grafting decision trees together and by making use of the hidden layers in the multi-layer perceptrons (explained in Section~\ref{sec:model}). Our evaluations show that the choice of classifier and the composition strategies that those classifiers afford yield varied, yet comparable results in a visual reference resolution task (Section~\ref{sec:exp}). Our analyses of the multi-layer perceptron model shows that the coefficients of the neurons in the hidden layers show properties that are similar to distributional embedding approaches to lexical semantics (Section~\ref{sec:analyses}), which we evaluate in Experiment 3. Finally, we conclude and explain our plans for future work.

\section{Related Work}

\newcite{Emerson2017} offer a comprehensive review and discussion on different approaches to compositionality including formal frameworks, tensor-based composition, and syntactic dependencies, to which we refer the reader. Their own model's application of composition uses distributional semantics and a probabilistic graphical model.

Recursive neural networks can arguably encode compositional processes directly, but it has been shown recently in \newcite{Lake2017} that they fail in a proof-of-concept neural machine translation task. The authors suggest a lack of systematicity in how these networks learn the composition process. Moreover, \newcite{Socher2014} reported successful application of composition using recursive neural models, but their approach made use of syntactic information from dependency parses; i.e., the composition was likely guided by the syntactic representation rather than the neural network. \newcite{yu2018mattnet} introduce MAttNet, which leverages recursive networks to obtain high results on a referring expression task using the same data we use for our experiments. In contrast with this work, our primary purpose is not to achieve state-of-the-art results, but rather a systematic check of composition strategies.

Comparable to their work and ours here is \newcite{Paperno2014} which attempted distributional composition using linguistic information similar to dependency parses. These approaches and tasks generally apply a \cta\ strategy of composition, where the final task is applied after combining meaning representations of the parts to first arrive at a single meaning representation of a sentence. Our work also compares to \newcite{Wu2019} which proposed a model that incorporates grounded and distributional information, which we explore in Experiment 3. 

We extend prior work to mitigate \wac's shortcomings by modeling several different \atc\ and \cta\ strategies made possible by the machinery of our chosen classifiers, and we use the resulting classifier coefficients as a possible way to bring \wac\ into a distributional space.\footnote{While we are not claiming that our approach is cognitively plausible, we take inspiration from \newcite{Barsalou2015} which notes that physical situations play central roles in establishing and using concepts--the \wac\ model is such a grounded model that considers physical context. Moreover, the \wac\ model is not an idealized account of concept representation; indeed, we take the vocabulary as our (very noisy) ontology and the classifiers learn probabilistic mappings which are far from idealized formal representations.}

\section{Data}

\begin{figure}[t]1
  \centering
      \includegraphics[width=0.35\textwidth]{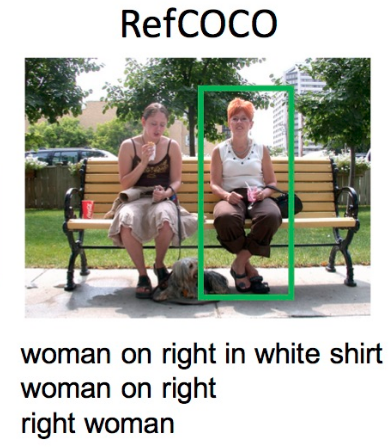}	
      \caption{Example of refCOCO image with three referring expressions (one on each line) made to an image region (see Footnote 3 for source).\label{fig:refcoco-example}}
\end{figure}

To give context to understanding our model and composition approaches, we first explain the data that we use in our experiments. We use the same dataset as described in \newcite{Schlangenetal2016}, the  “Microsoft Common Objects in Context” (MSCOCO) collection \cite{10.1007/978-3-319-10602-1_48}, which contains over 300k images with object segmentations, object labels, and image captions, augmented by \newcite{Mao2016} to add English referring expressions to image regions (i.e., refCOCO). The average length of the referring expressions is 8.3 tokens and the average number of image regions (which we treat as reference candidates in our experiments) is 8. We used the data access interface provided by \newcite{kazemzadeh2014referitgame}, which included a defined training/valiation/test split of the data.\footnote{\url{https://github.com/lichengunc/refer}} An example image, image region, and three referring expressions to that image region are depicted in Figure~\ref{fig:refcoco-example}. 

\section{Model}
\label{sec:model}

The \wac\ approach to lexical semantics is essentially a task-independent approach to predicting semantic appropriateness of words in physical contexts. The \textsc{wac} model pairs each word $w$ in its vocabulary $V$ with a classifier that maps the real-valued features $x$ of an entity $ent$ to a semantic appropriateness (i.e., class membership) score:

\vspace{-0.5cm}
\begin{center}
\begin{equation}
  \denotes{w}_{ent} = \lambda \mathbf{x}. p_w(\mathbf{x})
\label{eq:wac:intensobj}
\end{equation}
\end{center}

For example, to learn the connotative meaning of the word {\it red}, the low-level features (e.g., visual) of all objects referred to with the word {\it red} in a corpus of referring expressions are given as positive instances to a supervised learning classifier. Negative instances are randomly sampled from the complementary set of referring expressions (i.e., not containing the word \textit{red}). This results in a trained $\lambda \mathbf{x}.p_{red}(\mathbf{x})$, where $x$ is an object that can be applied to $red$ to determine class membership. 

\subsection{Approaches to Composition with WAC}

Traditionally, the \wac\ model has been applied using independent linear classifiers, such as logistic regression. In this paper, we expand upon the previous work by conducting experiments with a variety of classifiers, such as  multi-layer perceptrons (\mlp) and decision trees (\tree), in addition to logistic regression (\logreg). We use these classifiers so as to not make assumptions of linearity (i.e., \mlp) and to use the available machinery that \mlp s and \tree s afford us for exploring composition strategies. We explain below each approach and the methods of composition for each. It is important to note that in our explanations and discussions of composition, we are focusing on \emph{application} of the classifiers; all classifiers are trained as explained above (i.e., by pairing words with positive and negative examples). We leave more specific explanation of training to Section~\ref{sec:exp}. 

\subsubsection{Logistic Regression}

\paragraph{\logreg\ summed-predictions} The traditional approach to composition using \wac\ uses a \atc\ strategy where each \wac\ classifier in a referring expression is applied to an entity (e.g., a visually present object in a scene) which yields a probability (i.e., a score of fitness) for each word applied to each object. Following \newcite{Schlangenetal2016}, the resulting probabilities can be combined in several ways including summing, averaging, or multiplication (we opt for summing in our experiments) to produce a single overall expression-level fitness score for each object. This operation constitutes the composition of the referring expression into a single distribution (hence, \atc: the objects are \emph{applied} to each word classifier, then the resulting probabilities are composed into a single distribution over candidate objects). The object with the highest score is the hypothesized referred target.



\subsubsection{Multi-layer Perceptrons}

We explain in this section how we leverage \mlp s for three different approaches to composition. Unlike \logreg, the \mlp\ does not need to assume linearity, and its structure allows us greater flexibility for compositional techniques. 

\paragraph{\mlp\ summed-predictions}  For completeness and direct comparison to \logreg, we leverage \mlp s as we did with the \logreg\ \atc\ approach by simply using a \mlp\ (i.e., using a single hidden layer of 3 neurons) in place of a \logreg\ classifier for each word. Composition is applied after application by summing the resulting probabilities. 

\paragraph{Adjectives and Nouns} Following \newcite{baroni-zamparelli:2010:EMNLP}, we looked at adjective-noun pairs, which are two-word compositions that can identify how an adjective qualifies a noun. For example, instead of treating \textit{large}, \textit{green} and \textit{tree} separately in the phrase \textit{the large green tree next to the lake}, we use the adjectives \textit{large} and \textit{green} to modify the noun \textit{tree}. We explore two approaches to composing adj-noun pairs using \mlp s:

\textbf{\emph{\mlp\ adj-noun extended hidden layers}}: if there exist multiple adjectives for one noun, a separate adj-noun pair is created for each adjective preceding the noun. When predicting, the \mlp\ classifiers for each word in every adj-noun pair in the referring expression is merged together into one classifier by extending the hidden layer to include neurons from both the adjective and the noun in a single \mlp. The coefficients of the top layer (i.e., a binary sigmoid) of the original adj and noun \mlp s are averaged together to produce a single probability. The rest of the phrase is then composed normally using the traditional \wac\ methodology. This approach effectively leverages a connotative strategy to compose the adj-noun pairs, then a denotative strategy to compose the rest of the expression.\footnote{This differs from the \logreg\ \cta\ approach (i.e., where the coefficients were averaged) because in this case the coefficients in the hidden layer are not averaged, only the top layer coefficients are averaged.}

\textbf{\emph{\mlp\ adj-noun warm start}}: \mlp\ machinery also affords the use of the warm-start feature for the \mlp s. In this approach, we take the noun's classifier in the adj-noun pair which has already been trained, and then continue to train the classifier (i.e., using the warm-start functionality) with data that was used to train the adjective's classifier. This results in a single classifier that theoretically represents the entire adj-noun pair in a single classifier. This approach keeps the classifiers the same size, while leveraging a transfer-style learning approach to produce a adj-noun pair classifier that is composed of its constituent words. 

\paragraph{\mlp\ extended hidden layers} For this approach, we generalize the \mlp\ adj-noun extended hidden layers approach and apply it to the entire expression by concatenating the neurons in each word \mlp's hidden layer and averaging the coefficients of the top layer to form a single, composed \mlp\ that has a single hidden layer with nodes that can determine fitness between objects and all words in an expression.

\subsubsection{Decision Trees}

\begin{figure}[t]
  \centering
      \includegraphics[width=0.5\textwidth]{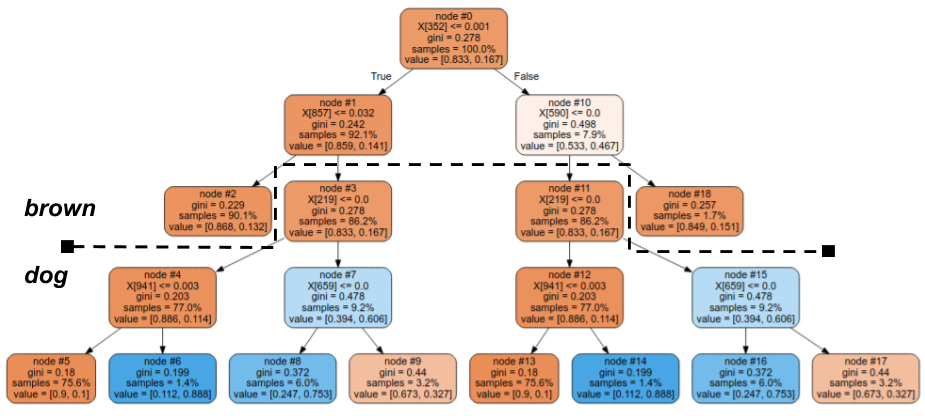}	
      \caption{Example of decision tree graft of \emph{brown} with \emph{dog}: the root node of \emph{dog} is grafted into the leaf nodes that would output a “true” classification \emph{brown}. This composes the noun-phrase “brown dog” (the dog classifier augments the brown classifier).\label{fig:tree-graft}}
\end{figure}

We opted to apply \tree s because of their readable internal structure and how the branching system lends itself to intuitive composition strategies. 

\paragraph{\tree\ summed-predictions}  For completeness and direct comparison, we leverage \tree s as we did with the \logreg\ and \mlp\ \atc\ approaches by simply using a \tree\ classifier for each word. Composition happens after application by summing the resulting predicted probabilities. 

\paragraph{\tree\ grafting} In this approach to composition, we leverage the \tree\ representation by training the classifier as usual (i.e., independently), but then to compose the classifiers into a single classifier, we ``graft'' the trees of each classifier $w_i$ in a referring expression to the two most probable ``true''  leaf nodes of the classifier $w_{i-1}$, beginning with $w_1$ as the starting tree with the root node. This grafting process of two words is depicted in Figure~\ref{fig:tree-graft} where the root node of the decision tree for the word \emph{dog} is grafted into the true leaf nodes of the decision tree classifier for the word \emph{brown}, thus allowing each word to make a contribution of the final decision of expression-level fitness to objects. This grafting process is repeated for each word in the referring expression, thus allowing us to apply then compose an entire phrase.

\subsection{Composing Relational Expressions}

For this final model, we follow and extend \newcite{Kennington2015a}'s approach to relational expressions (e.g., containing prepositions such as \emph{next to} or \emph{above}, etc.) by expressing those relational phrases ($r$) as a fitness for pairs of objects (i.e., $R_1$ and $R_2$; whereas for non-relational words, \wac\ only learns fitness scores for single objects):

\vspace{-0.2cm}
{\small
\begin{equation}
  \denotes{r}^W = P(R_1,R_2|r)
  \label{eq:rel}
\end{equation}
}
\vspace{-0.2cm}

However, in prior work, the two candidate objects were known. In our model, we only assume that the final referred target object is known (i.e., $R_1$), but the relative object is not (i.e., $R_2$). To handle this, we interpret relational phrases $r$  containing prepositional words as ``transitions'' from one noun phrase ($NP_1$) to the relative noun phrase ($NP_2$). For example, in the phrase \emph{the woman to the right of the tree}, we first identify the relational phrase(s) (in this case \emph{right of}), and we use this relational phrase to make the most likely transition from $NP_1$ (i.e., \emph{the woman}) to $NP_2$ (i.e., \emph{the tree}). We do this by using a learned \wac\ model (i.e., trained on single objects using any of the approaches explained above) and apply $NP_2$ to all candidate objects in the scene. This produces a distribution over all objects (i.e., a partially observable $NP_2$); we use the resulting argmax of that distribution to arrive at the $R_2$ that was referred to by $NP_2$. With $R_1$ and $R_2$, we train $r$ by taking the difference in features (i.e., simple vector subtraction) between the feature vector for $R_1$ and $R_2$ resulting in a feature vector for $r$ that is the same size as for all other \wac\ classifiers. During application, we find the most likely pair by applying $NP_1$ and $NP_2$ to all objects, then $r$ to all pairs of objects, and force identity on the $NP_1$ distribution and candidate $R_1$, as well as the $NP_2$ distribution and the candidate $R_2$, forming a trellis-like structure. The combined (i.e., product) of probabilities for $NP_1$, $r$, and $NP_2$ result in a final distribution; we take the argmax probability and the resulting $R_1$ object as the target referred object.

\section{Experiments}
\label{sec:exp}

\subsection{Experiment 1: Simple Expressions}

In this experiment, we evaluate the performance of our approaches to composition of simple (i.e., containing no relational words) referring expression resolution using the refCOCO data. 

\paragraph{Task \& Procedure} The task is reference resolution to objects depicted in static images. For example, the referring expression \textit{the woman in red sitting on the left} would require composition of each word (with the exception of the quantifier \emph{the} which signals that the referring expression should only identify a single entity). We task the \wac\ model (and varying compositional approaches, as explained above) to produce a ``distribution'' (i.e., ranked scores) of each object in an image. We only considered referring expressions that did not contain words included in this list: \emph{below, above, between, not, behind, under, underneath, front of, right of, left of, ontop of, next to, middle of} (as was done in \newcite{Schlangenetal2016}) resulting in simpler referring expressions. This resulted in 106,336 training instances and 9,304 test instances (i.e., individual referring expressions and the corresponding image with multiple object regions). In Experiment 2 we consider all referring expressions, including those with relations.

Training the \wac\ model follows prior work where each word in the refCOCO training data is trained on all instances where that word was used to refer to an object and 5 randomly chosen negative examples sampled from the training data where that particular word was not used in a referring expression (i.e., a 5-to-1 negative to positive example ratio). For each object in the images, following \newcite{Schlangenetal2016}, we used the image region information from the annotated data and extract that region as a separate image that we then pass through GoogLeNet \cite{Szegedy2015}, a convolutional neural network that was trained on data from the Large Scale Visual Recognition Challenge 2014 (ILSVRC2014) from the ImageNet corpus \cite{JiaDeng2009}. That is, GoogLeNet is optimized to recognize single objects within an image, which it can do effectively with our extracted image regions. This results in a vector representation of the object (i.e., region) with 1024 dimensions (i.e., we use the layer directly below the predictions layer).\footnote{We used the development set to evaluate other existing neural networks trained on the ImageNet data, including approaches more recent and with better performance on the task than GoogLeNet, but we found that GoogLeNet worked better for our task.} We concatenate to this vector 7 additional features that give information about the image region resulting in a vector of 1031 dimensions: the (relative to the full image from which it was extracted) coordinates of two corners, its (relative) area, distance to the center, and orientation of the image. 

To ensure that our \wac\ models were made up of reliable classifiers, we threw out words that had 4 or fewer positive training examples, resulting in an English vocabulary of 2,349. For training the \logreg\ \wac\ model, we used $L1$ normalization. For training the \tree\ \wac, we used the GINI splitting criteria with a maximum depth of 2. For training the \mlp\ \wac\ model, we used a multi-layer perceptron with a single hidden layer of 3 neurons using the tanh activation function, the top layer is a binary sigmoid, trained using the adam solver (alpha value of 0.1) for 2000 maximum epochs. In all cases, the best hyper parameters were found using the development data.

\paragraph{Metrics} The metric we use for this task is \emph{accuracy} that the highest scoring object in an image of candidate objects as ranked by our model matches the annotated target object referred to by the referring expression. For example, in Figure~\ref{fig:refcoco-example}, there are several annotated objects including the two people, the trees, car, ground, etc., but the referring expression \emph{woman on the right in white shirt} should uniquely identify the target object in the annotated image region. Our target is 0.64, the result reported in \newcite{Schlangenetal2016}.

\paragraph{Results}

\begin{table}
{\small
\centering
 \begin{tabular}{|l|c|c|}
\hline
 \textbf{model} & \textbf{Exp. 1 acc} & \textbf{Exp. 2 acc} \\
\hline
\logreg\ summed-preds & \textbf{0.64} & 0.62 \\
\mlp\ summed-preds & 0.63 & 0.61 \\
\mlp\ adj-noun ext. hidden & 0.62 & 0.60 \\
\mlp\ adj-noun warm-start & 0.62 & 0.61 \\
\mlp\ ext. hidden & 0.6  & 0.58\\
\mlp\ relational & 0.62  & \textbf{0.63} \\
\tree\ summed-preds & 0.55 & 0.54\\
\tree\ graft & 0.39 & 0.27\\
\hline
\end{tabular}
\caption{Experiment 1 \& 2 results: accuracy scores for the models and composition approaches using the refCOCO data.}
\label{tab:exp_results}
}
\end{table}

The results for this experiment are in the \textbf{Exp.\ 1 acc} column of Table~\ref{tab:exp_results}.\footnote{\newcite{yu2018mattnet} yielded far better results than we show here--our goal is to explore compositionality directly using a more transparent model.} We see that in all cases, the \atc\ \emph{summed-predictions} for all three classifiers yield respectable performance (the first row verifies results reported in \newcite{Schlangenetal2016} for this task). When considering \cta\ approaches, the results are more nuanced: for the \mlp\ adj-noun approaches, we see similar results when the hidden layers are extended and when we use warm-start. This is a positive result in that composing adj-noun pairs together using warm-start is theoretically appealing because a single classifier can perform the work of two, though it needs additional training data to perform that function, whereas extending the hidden layers requires no additional retraining, and the compose-then-apply nature of it makes it more appealing than simply summing the predictions of each word applied to the objects, as has been done in prior work. Unfortunately, when extending the hidden layers to contain the neurons from the \mlp\ classifiers for all words in the expression, the results take a hit when compared to the \atc\ summed predictions approach. The \tree\ classifiers did not perform well in this particular task (both the \atc\ summed predictions and the grafting) which is somewhat surprising, though as can be seen in Figure~\ref{fig:tree-graft}, by limiting the depth, the \tree\ classifiers were unable to make use of the nuance in the individual features. 

\subsection{Experiment 2: Referring Expressions with Relations}

For this experiment, we apply our approaches of composition to all of the refCOCO data, including cases where relations are present. 

\paragraph{Task \& Procedure} The task and procedure for this experiment are similar to that of Experiment 1 with the important difference that we consider all of the training and test instances (120,266 and 9,914, respectively) for training; we do not ignore referring expressions that have relations, for example \textit{the woman in red sitting on the left next to the tree}.

\paragraph{Metrics} The metrics for this experiment are the same as Experiment 1: accuracy that the highest scoring object as ranked by our model matches the annotated target object referred to by the referring expression.

\paragraph{Results} Table~\ref{tab:exp_results}, column \textbf{Exp.\ 2 acc} contains the results for this experiment. As expected, nearly all approaches took a hit because they were unable to handle the composition of multiple noun phrases and learn the semantics of relational words. The best performing model, as expected, is the \mlp\ relational model which applies the extended hidden layer to individual noun phrases, but composes them using the relational word using summed predictions (the results are statistically significant compared to the \logreg\ summed-preds row result for Exp 2).  The final relational model, which uses the extended hidden approach for each noun phrase, applies more principled \cta\ composition and works with relational expressions without requiring annotation of the relative object.

\subsection{Analyses}
\label{sec:analyses}

We end this experiment by offering some analysis on what the \wac\ classifiers are learning and compare some of the composition strategies with each other. We put focus on the \mlp\ classifier since it produced the best results and allows more possible composition strategies. 

\begin{figure}[h]
  \centering
      \includegraphics[width=0.3\textwidth]{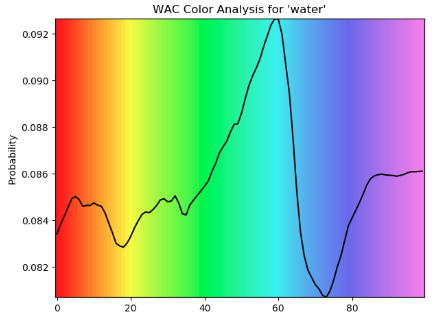}	
      \caption{Probabilities returned by passing color images through the \emph{water} classifier from the red to the violet range. \label{fig:wac-water}}
\end{figure}

\paragraph{Individual Classifiers} Similar to \newcite{Kennington2015a}, we looked at classifiers that, we assume, learned features relating to colors. To check this, we ranged over a wide range of colors, producing an image for each color, and passed those through the GoogLeNet to arrive at a representation for each color that our model could use. Our analyses show that the classifiers learned prototypical colors for objects as well as color terms. For example, we passed all colors through the \emph{water} classifier and plotted the resulting probabilities, resulting in Figure~\ref{fig:wac-water}. This shows that individual classifiers can pick up on color information where an important visual property is its color (i.e., \emph{water}), though that information is not apparent in the underlying representation (i.e., the GoogLeNet vector). 

This is an important result with useful implications: this kind of transfer learning using GoogLeNet (or any other model) trained on the ImageNet data extends the original limited ability of those networks by using the entire vocabulary to identify not only object types, but also attributes (e.g., colors, sizes, spatial placements, etc.). Importantly for interactive dialogue with robots, the \wac\ approach allows new vocabulary to be added word-by-word without requiring retraining of the entire underlying network.

\paragraph{Semantic Clusters} To see if \wac\ could also yield semantic clusters, we applied a t-distributed Stochastic Neighbor Embedding (TSNE) \cite{Maaten2008} to the hidden layers of the \mlp\ classifiers (i.e., 3021 dimensions, as each hidden layer had 3 neurons, each neuron had coefficients for 1007 features), mapping the coefficient vectors to 2 dimensions, then applied the Density-based Spatial Clustering of Applications with Noise (dbscan; $eps=0.7$) \cite{Schubert2017} to cluster the TSNE result, which is depicted in Figure~\ref{fig:dbscan-wac}.\footnote{We found that logistic regression coefficients did not result in any meaningful clusters.} The figure shows a large, central cluster with more informative clusters closer to the edges. The following are several noteworthy clusters:

\begin{figure}[h]
  \centering
      \includegraphics[width=0.51\textwidth]{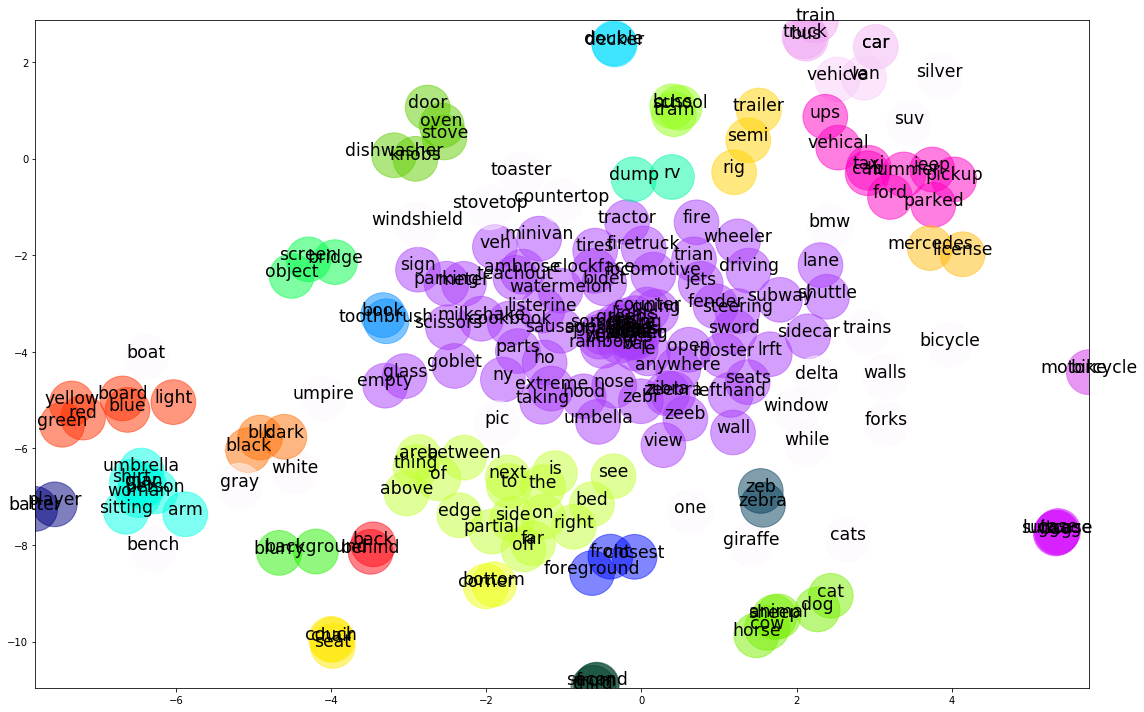}	
      \caption{Cluster proposals of \mlp\ classifier hidden layer node coefficients. \label{fig:dbscan-wac}}
\end{figure}

{\small
\vspace{-0.5cm}
\begin{itemize}
\setlength\itemsep{-0.5em}
 \item \emph{yellow, red, green, blue, light, board}
 \item \emph{area, between, of, above, edge, next to, right}
 \item \emph{door, oven, stove, dishwasher, knobs}
 \item \emph{trailer, semi, rig, car, vehicle, van, ups, suv, taxi}
 \item \emph{cat, dog, horse, cow, sheep, animal}
\end{itemize}
}

This analysis informs us that using the \mlp\ classifier is not only more satisfying for composition approaches, but it can also yield coefficients that can be used as vectors for word embeddings. Our final experiment tests this hypothesis.


\subsection{Experiment 3: WAC Coefficients for Semantic Similarity}

In this experiment, we evaluate the vectors that are derived from the coefficients of the \mlp\ variant of \wac\ in a semantic similarity task. We follow a similar approach to \newcite{kottur2016visual} by creating visually grounded word embeddings, but we make use of the WAC model and we create our own dataset for this experiment.

\paragraph{Task \& Procedure} We evaluated the \wac\ coefficient vectors using the semantic similarity tasks WordSim-353 \cite{Finkelstein2002} and SimLex-999 \cite{Hill2015} by comparing the \wac\ vectors with GloVe embeddings \cite{Pennington2014} (6B, 200d), and a concatenation of the vectors for each word in the two models. To arrive at a model for \wac\ with a large enough vocabulary, we retrieved 100 images (i.e., using Google Image Search) for each of 30,000 common English words and trained the \mlp\ classifiers as described in Experiment 1 above using 100 images for each word. 

\paragraph{Metrics} For semantic similarity, we report a Spearman Correlation between the cosine similarities of the \wac\ coefficient vector, the GloVe embeddings, and the combination of the two models. High numbers denote better scores. 

\paragraph{Results} The results of the Spearman Correlation is shown in Table~\ref{tab:exp3_results}. The results show that, when combined with embeddings trained on large amounts of text such as GloVe, the \wac\ coefficient vectors can contribute useful information derived from the visual features that the GloVe embeddings don't have. We conjecture that \wac\ does not work well on its own because the words are trained independently, and the \wac\ model assumes that words trained on visual representations are concrete, whereas many words are abstract, the semantics of which is learned in lexical context, such as distributional embeddings. These results show that \wac\ is a potential model for bridging grounded and distributional approaches to lexical semantics, which we will explore further in future work. 

\begin{table}
{\small
\centering
 \begin{tabular}{|l|c|c|}
\hline
 \textbf{model} & \textbf{WordSim-353} & \textbf{SimLex-999} \\
\hline 
\wac\           & 0.485 & 0.157 \\
GloVe           & 0.630  & \textbf{0.339} \\
combined        & \textbf{0.707} & 0.326 \\
\hline
\end{tabular}
\caption{Experiment 3 results: Spearman Correlation between the cosine similarities of \wac, GloVe, and the two combined.}
\label{tab:exp3_results}
}
\end{table}


\section{Conclusions}

In this paper, we explored using \logreg, \mlp, and \tree\ for composing the words-as-classifiers approach to grounded, lexical semantics. We evaluated several methods including \atc\ where the classifiers are applied to objects, then the resulting probabilities are summed, \cta\ where the composition was applied first (i.e., either by extending hidden layers or using warm start in \mlp, or grafting \tree). Our results show that classifiers affect results, and the kinds of composition that can be accomplished varies depending on the classifier.  We concur with \newcite{Baroni2019a} that more exploration needs to be done in this area to learn the systematic functional applications of composition of language; our results are an important step in that direction. 

Furthermore, we showed that the \wac\ model has properties that lends itself for a straightforward embedding by using the classifier coefficients. We showed in Experiment 3 that the embeddings show promise in a simple word similarity task. We are in the process of evaluating the \wac\ embeddings on other common tasks and analyzing the efficacy of \wac\ embeddings by combining them with other approaches to semantics. 

Combined with prior work, the results reported here have implications for tasks and settings for speech interaction: \wac\ can be trained with minimal training data, used in dialogue systems where words ground to multiple modalities (e.g., vision, proprioperception, predicted emotions, etc.), can be composed incrementally, and the coefficients of the classifiers can be unified with other distributional representations. 

For future work, we will explore how \wac\ can be coupled with formal and distributional semantic representations to better exploit and integrate knowledge from multiple modalities for a more holistic representation of semantics. We are also evaluating \wac\ in an interactive language learning task with two different robot platforms. 



\bibliographystyle{acl_natbib}
\bibliography{refs}

\end{document}